\documentclass[journal,twoside,web]{ieeecolor}
\usepackage{generic}
\usepackage{cite}
\makeatletter
\let\NAT@parse\undefined
\makeatother
\usepackage[colorlinks=true,citecolor=blue,linkcolor=blue,urlcolor=blue,hypertexnames=false]{hyperref}
\usepackage{amsmath,amssymb,amsfonts}
\usepackage{booktabs}
\usepackage{algorithmic}
\usepackage{graphicx}
\usepackage{textcomp}
\usepackage{multirow}
\usepackage{amssymb}
\usepackage{pifont}
\def\BibTeX{{\rm B\kern-.05em{\sc i\kern-.025em b}\kern-.08em
    T\kern-.1667em\lower.7ex\hbox{E}\kern-.125emX}}
\begin{document}
\title{Maximizing T2-Only Prostate Cancer Localization from Expected Diffusion Weighted Imaging}
\author{Weixi Yi, Yipei Wang, Wen Yan, Hanyuan Zhang, Natasha Thorley, Alexander Ng, Shonit Punwani, Fernando Bianco, Mark Emberton, Veeru Kasivisvanathan, Dean C. Barratt, Shaheer U. Saeed, Yipeng Hu
\thanks{This work is supported by the International Alliance for Cancer Early Detection, an alliance between Cancer Research UK [EDDAPA-2024/100014; C73666/A31378], Canary Center at Stanford University, the University of Cambridge, OHSU Knight Cancer Institute, University College London and the University of Manchester. The authors acknowledge the use of resources provided by the Isambard-AI National AI Research Resource (AIRR). Isambard-AI is operated by the University of Bristol and is funded by the UK Government’s Department for Science, Innovation and Technology (DSIT) via UK Research and Innovation; and the Science and Technology Facilities Council [ST/AIRR/I-A-I/1023].}
\thanks{Weixi Yi, Yipei Wang, Wen Yan, Hanyuan Zhang, Dean C. Barratt, Shaheer U. Saeed, and Yipeng Hu are with the UCL Hawkes Institute and the Department of Medical Physics and Biomedical Engineering, University College London, London, UK.}
\thanks{Shaheer U. Saeed is also with the Centre for Bioengineering, School of Engineering and Materials Science and Digital Environment Research Institute, Queen Mary University of London, London, UK.}
\thanks{Natasha Thorley and Shonit Punwani are with the Centre for Medical Imaging, University College London, London, UK.}
\thanks{Alexander Ng, Mark Emberton and Veeru Kasivisvanathan are with the Division of Surgery \& Interventional Science, University College London, London, UK.}
\thanks{Fernando Bianco is with Urological Research Network, Miami Lakes, Florida, USA}
\thanks{Corresponding author: Weixi Yi (email: weixi.yi.22@ucl.ac.uk)}}

\maketitle

\begin{abstract}
Multiparametric MRI is increasingly recommended as a first-line noninvasive approach to detect and localize prostate cancer, requiring at minimum diffusion-weighted (DWI) and T2-weighted (T2w) MR sequences. Early machine learning attempts using only T2w images have shown promising diagnostic performance in segmenting radiologist-annotated lesions. Such uni-modal T2-only approaches deliver substantial clinical benefits by reducing costs and expertise required to acquire other sequences. This work investigates an arguably more challenging application using only T2w at inference, but to localize individual cancers based on independent histopathology labels. We formulate DWI images as a latent modality (readily available during training) to classify cancer presence at local Barzell zones, given only T2w images as input. In the resulting expectation-maximization algorithm, a latent modality generator (implemented using a flow matching-based generative model) approximates the latent DWI image posterior distribution in the E-steps, while in M-steps a cancer localizer is simultaneously optimized with the generative model to maximize the expected likelihood of cancer presence. The proposed approach provides a novel theoretical framework for learning from a privileged DWI modality, yielding superior cancer localization performance compared to approaches that lack training DWI images or existing frameworks for privileged learning and incomplete modalities. The proposed T2-only methods perform competitively or better than baseline methods using multiple input sequences (e.g., improving the patient-level F1 score by 14.4\% and zone-level QWK by 5.3\% over the T2w+DWI baseline). We present quantitative evaluations using internal and external datasets from 4,133 prostate cancer patients with histopathology-verified labels. The code is available at \url{https://github.com/wxyi057/T2-GEM}.
\end{abstract}


\section{Introduction}

\begin{figure}[t]
\centering
\includegraphics[width=\columnwidth]{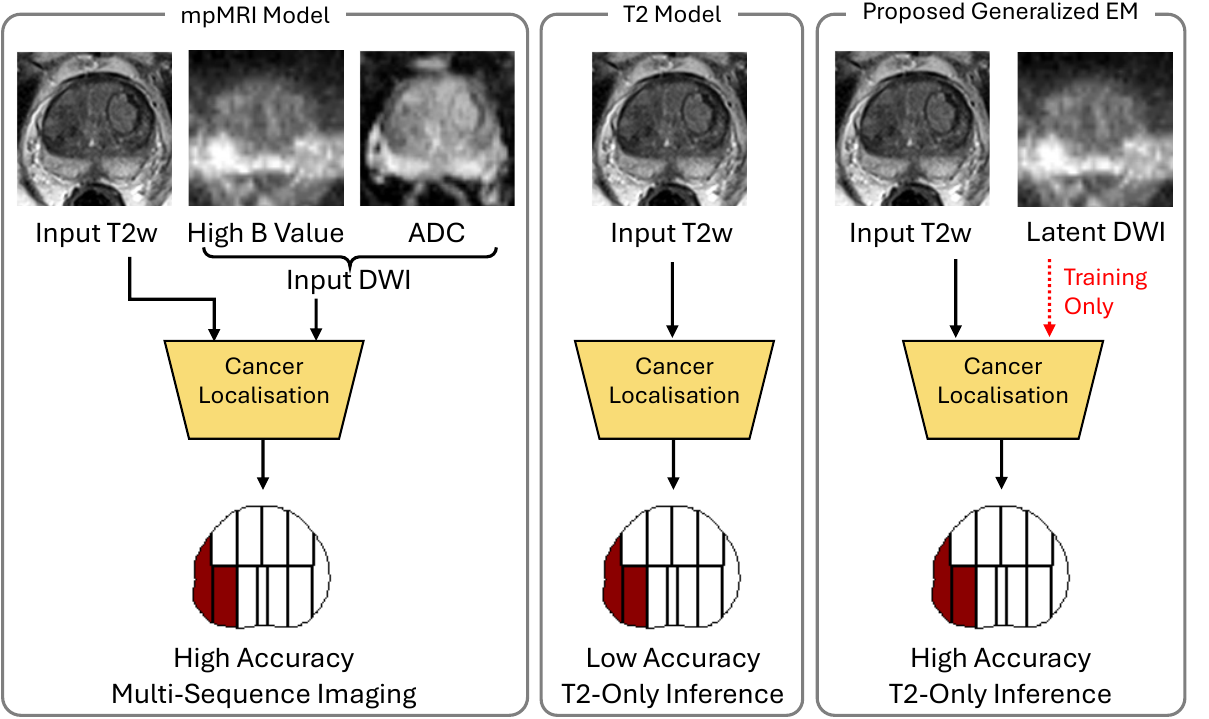}
\caption{Prostate cancer localization paradigms. Unlike standard mpMRI requiring multi-sequence inputs, our proposed Generalized EM leverages DWI strictly as a latent modality during training. This enables high-accuracy inference using only T2w images, overcoming the limitations of standard T2 models. Red denotes ISUP$\ge$3.} \label{fig1}
\end{figure}

Multiparametric MRI (mpMRI), combining T2-weighted (T2w) imaging with diffusion-weighted imaging (DWI; from which apparent diffusion coefficient, ADC, maps are derived) and dynamic contrast enhanced (DCE) imaging, has become the clinical standard for prostate cancer localization \cite{turkbey2019prostate}. Growing approaches have been proposed to use modern machine learning to automatically detect prostate cancer on mpMRI for patient-level diagnosis, e.g. \cite{cai2024fully,xu2024poisson} and cancer localization, such as pixel-level segmentation e.g. \cite{lee2025prostate,yan2024combiner} or zone-level classification e.g. \cite{wu2024ai}. These studies have been validated using radiologist labels e.g. \cite{yi2024t2} or histopathology-verified labels e.g. \cite{alfano2022prostate,schelb2019classification,lee2025prostate}. 

A prior proof-of-concept study \cite{yi2024t2} demonstrated that deep learning models relying solely on T2w modalities can achieve performance comparable to mpMRI approaches in the segmentation of radiologist-annotated lesions, offering substantial advantages regarding imaging speed and accessibility. However, a reliance on radiologist annotations bounds model performance to human visual sensitivity. This sensitivity is subject to variability and errors, which may render the approach suboptimal for accurate interventions.

In addition to mimicking radiological perception, recent studies have also explored directly predicting histopathology. Cancer locations are labelled using sparse, biologically grounded evidence derived from systematic saturation biopsies, as opposed to dense and subjective lesion contours. To the best of our knowledge, this work represents a novel application of the latent-variable modeling of T2-only localization of prostate histopathology cancers, and its internal and external evaluation, compared to existing literature (including the previous study \cite{yi2024t2}).

Radiological evidence supports the use of DWI and additional DCE sequences \cite{tamada2021comparison}, which may contribute to our proposed T2-only approach perceived as counter-intuitive. Therefore, we discuss the potentials and feasibility in using T2w sequences only, differentiating radiologist examination from our proposed machine learning approach.
First, 20\% radiologist-missed cancers may be detected by machine learning models \cite{seetharaman2021automated,sun2023multicenter}, a subset of which may be solely predictable from T2w images. This potential is related to the inherent limitation of human radiologist perceptive ability, with which positive signals from T2w images may not be reliably visible to them.
Second, machine learning models may require additional inter-modality registration to compensate the difference between T2w and DWI images, e.g. due to organ movement and motion artifacts \cite{yang2022cross}, for efficient learning using multi-modality image inputs - especially for cancer localization tasks (as opposed to patient-level classification). This remains a challenging image fusion task, one that experienced observers may less rely on. 
Third, individual radiologists are more likely to be biased, due to diverse clinical experience and training, with varying image quality and protocols \cite{tang2025impact}. Machine learning models trained with histopathology labels are independent of such radiologist biases. 
Finally, DWI can still be utilised for training-only purposes, for example, by learning the correlation between T2w and DWI in the large amount of existing training data, which can then be leveraged at inference without needing DWI at test time.

Compared with our previous work \cite{yi2024t2}, the present study differs in four main respects. First, the methodology changes from an empirical bi-level meta-learning framework, which aligns a modality translator and a predictor, to a probabilistic latent modality formulation with a rigorous mathematical basis, solved via a generalized expectation-maximization algorithm. A flow-matching generator approximates the posterior distribution of the missing DWI modality, and the cancer localizer is then trained by marginalizing over this uncertainty. Second, the task changes from 2D pixel-wise lesion segmentation on cropped slices \cite{yi2024t2} to 3D zone-level localization \cite{wu2024ai}, which predicts the ordinal degree of malignancy for each predefined Barzell zone. Third, the supervision changes from radiologist-annotated lesion contours \cite{yi2024t2, yan2024combiner} to histopathology-based labels derived from systematic saturation biopsies. Fourth, the evaluation changes from segmentation metrics to a clinical assessment in which risk and grade stratification are examined across multiple anatomical levels (the Barzell zone, the peripheral or transition zone, and the patient level) and ISUP thresholds \cite{ahmed2017diagnostic}, using ordinal and discriminative metrics together with clinical operating points.

These motivated this work, with the following contributions, 1) a novel clinical application using only T2w images for prostate cancer localization at inference; 2) a novel probabilistic expectation-maximization framework that predicts cancer location on a single image modality, benefiting from an additional latent image modality in training; 3) an efficient and practical learning algorithm, with open-source implementation, to simultaneously optimize a cancer localizer and a latent modality generator; 4) a rigorous validation based on two clinical data sets with histopathology-verified cancer labels, presenting a set of quantitative comparison and ablation studies.

\section{Related Work}
\subsection{Automated Diagnosis and Localization on Prostate MRI}
At the patient level, deep learning models have achieved radiologist-level performance under PI-RADS–guided clinical assessment for prostate cancer risk stratification on mpMRI\cite{cai2024fully,xu2024poisson,hiremath2021integrated,shao2022patient,weisser2024weakly,wu2025automated,rodrigues2025improving}. For example, Cai et al.\cite{cai2024fully} proposed a fully automated pipeline for clinically significant prostate cancer (csPCa) detection on mpMRI, reporting diagnostic accuracy comparable to expert radiological reads. Xu et al.\cite{xu2024poisson} introduced a Poisson Ordinal Network to estimate Gleason score from mpMRI by explicitly modeling the ordinal nature of histologic grade. These patient-level models are well suited for triage by summarizing global imaging cues into a subject-level risk score, yet follow-up patient care options including targeted biopsy and focal therapy require the location of positive lesions. To support procedural guidance, many approaches often formulate this as voxel-wise segmentation or detection \cite{pellicer2022deep,lee2025mri,lin2024evaluation,fassia2024deep,li2024deep,yan2024combiner,cao2019joint,yi2024t2}. For instance, Cao et al.\cite{cao2019joint} introduced FocalNet for joint cancer detection and Gleason score prediction directly on mpMRI. Yan et al.\cite{yan2024combiner} proposed Combiner and HyperCombiner networks, which systematically investigate rules for fusing information from mpMRI to enhance segmentation accuracy. While offering fine-grained localization ideal for computer-aided diagnosis workflows, these methods heavily rely on pixel-level annotations (e.g., lesion masks), which are labor-intensive to acquire and prone to inter-observer variability among radiologists. Structured zone-level localization aligning with clinical reporting standards. Instead of predicting arbitrary lesion shapes, these methods classify cancer presence within predefined anatomical zones (e.g., Barzell zones) \cite{zeevi2023reliable,wu2024ai}. Such template-based systems are particularly relevant for biopsy planning and pathology reporting. For example, Wu et al.\cite{wu2024ai} developed an AI-assisted system to classify radiologist-defined regions of interest on mpMRI, streamlining the diagnostic workflow.

Most existing methods are trained and/or evaluated using radiology-derived labels, including PI-RADS assessments or radiologist-delineated ROIs \cite{lee2025mri,lin2024evaluation,fassia2024deep,li2024deep,yan2024combiner,yi2024t2}. While convenient and widely available, such annotations are inherently indirect surrogates of tumor extent and are subject to inter-reader variability. More recent studies have moved toward histopathology-confirmed labels from biopsy/TPM or prostatectomy whole-mount histology to obtain more reliable endpoints and stricter evaluation protocols \cite{zeevi2023reliable,wu2024ai,cai2024fully,xu2024poisson,hiremath2021integrated,shao2022patient,weisser2024weakly,wu2025automated}.

Another key axis is imaging modality availability. mpMRI (T2w, DWI, and DCE) is widely considered the most comprehensive protocol for lesion detection, localization, and risk stratification \cite{turkbey2019prostate}. However, longer acquisition times, contrast-agent-related risks, and higher costs have led many centers to routinely omit DCE, while still maintaining comparable diagnostic performance\cite{ng2025biparametric}. This shift has driven automated methods to reduce dependence on full multi-sequence inputs and explore deployment using T2w imaging alone. Jin et al. \cite{jin2024t2} showed that T2-only models can achieve effective cancer detection and Gleason grading on multi-centre data, indicating that anatomical imaging retains grade-related cues. However, T2w alone may fail to capture functional information from DWI, especially for subtle lesions and pathology-coupled endpoints. To bridge this gap while preserving T2-only inference, recent work incorporates multimodal supervision during training and distils complementary knowledge into T2-only models; for instance, Yi et al. \cite{yi2024t2} used meta-learning to transfer information from T2w and DWI, achieving strong performance on radiologist-annotated lesion segmentation with T2-only inference.

Despite these advances, prior work rarely studies pathology-grounded zone-level localization under T2-only inference. 

\subsection{Handling Reduced Modalities and Training-Privileged Information}

The first approach focuses on learning shared representations resilient to data absence. Classical approaches like HeMIS \cite{havaei2016hemis} compute statistical moments of available embeddings, while recent advances have extended this to hierarchical multi-modal variational models (e.g., H-UVED \cite{dorent2019hetero}), region-aware fusion (e.g., RFNet \cite{ding2021rfnet}), transformer-based unified representations (e.g., mmFormer \cite{zhang2022mmformer}) and adaptive expert-based fusion (e.g., SimMLM \cite{li2025simmlm}). Training strategies such as ModDrop \cite{neverova2015moddrop} further enforce robustness by randomly masking modalities to simulate inference-time unavailability. However, these methods inherently treat missing data as sporadic noise or redundant views, leading to suboptimal performance when a critical information source like DWI is systematically and entirely unavailable at inference. 

The second approach, learning using privileged information (LUPI), transfers knowledge from a multi-modal teacher to a uni-modal student. This has evolved from Vapnik's Support Vector Machines \cite{vapnik2009new} to deep distillation frameworks: KD-Net \cite{hu2020knowledge} aligns intermediate features, while recent works leverage prototype-based privileged knowledge (e.g., ProtoKD \cite{wang2023prototype}) to enhance representation transfer. 

The third approach, modality completion, explicitly synthesizes the missing sequences using generative models to serve as inputs for downstream tasks \cite{yi2024t2,wang2024b,zhou2020hi}. Nevertheless, a limitation in such approaches is the optimization decoupling between synthesis and diagnosis: generators are typically trained to minimize pixel-level reconstruction error rather than to maximize diagnostic utility.

In this work, we propose a unified probabilistic framework that treats the missing DWI not as a fixed target for separate reconstruction, but as a latent variable. 

\section{Methods}

\begin{figure*}[t]
\centering
\includegraphics[scale=0.75]{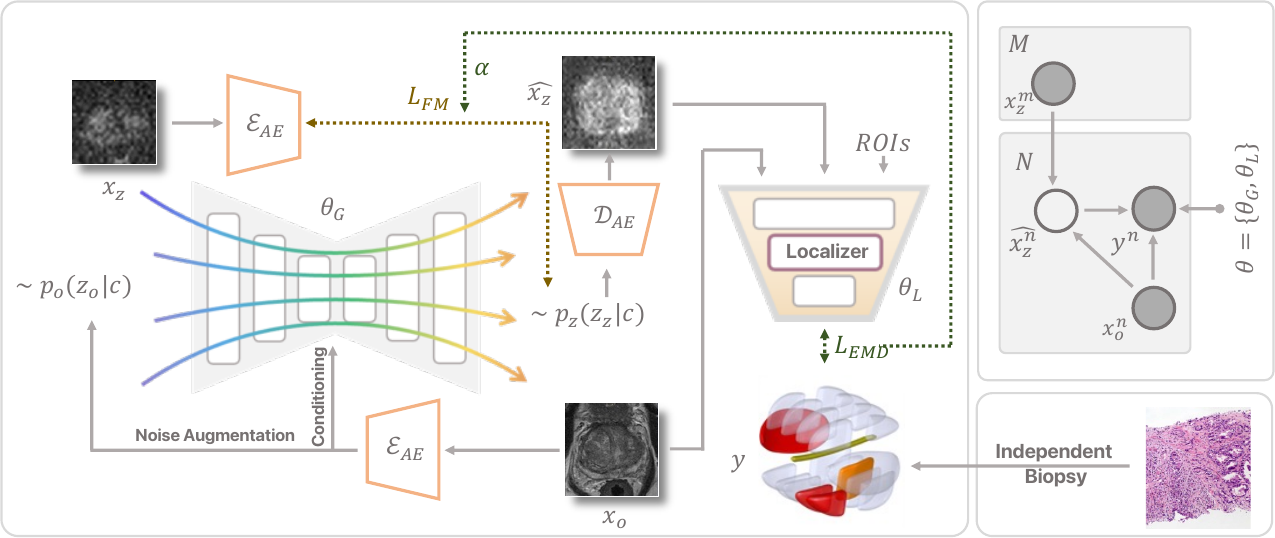}
\caption{The left shows the proposed GEM framework with two components: a latent modality generator $f_G$ generating latent images $\hat{x}_z$ from observed images $x_o$, and a cancer localizer $f_L$ localizing lesions using $x_o$ and $\hat{x}_z$. Biopsy-derived Barzell-zone labels $y$ are colored orange and red for ISUP grades 2 and $\ge$ 3, with others labeled negative. The right shows GEM’s graphical model: the latent variable $\hat{x}_z$ is generated from $x_o$ and $\theta$ (E-step), while $\theta$ is updated using $\hat{x}_z$ and data $\mathcal{T}$ (M-step). $r$ is omitted for clarity.}
\label{fig2}
\end{figure*}

As shown in Fig.~\ref{fig2}, we formulate the T2-only prostate cancer localization problem within a probabilistic Generalized Expectation-Maximization (GEM) framework. Let $x_o \in X_o$ denote the observed T2w image, and $x_z \in X_z$ denote the DWI image, the latter of which is treated as a latent variable available only during training. The localization target is the ordinal histopathology grade $y=\{y_r\}_{r\in R}$ for a set of predefined anatomical regions (here, Barzell zones \cite{ahmed2017diagnostic}) $R$.

Our framework maximizes the expected log-likelihood of the joint distribution through two coupled components: a Latent Modality Generator $f_G$ that approximates the posterior of the missing modality, and a Cancer Localizer $f_L$ that predicts malignancy using both observed and latent representations.

\subsection{Latent Modality Generation}

The latent modality generator $f_G$ leverages the flow matching model \cite{lipman2022flow,schusterbauer2024fmboost} to describe the generation process from observed images $X_o$ to latent images $X_z$ as a straight-line conditional probability path between two distributions within the framework of optimal transport. Compared to diffusion models \cite{ho2020denoising,rombach2022high}, this method not only significantly reduces numerical simulation errors but also requires only a few ODE solving steps to complete the simulation efficiently. The latent modality generator is defined as
$ f_G(\cdot; \theta_G): X_o \to X_z $.

To efficiently model the non-linear mapping between anatomical and functional modalities, we operate within a compressed, continuous latent space. We employ a single AutoencoderKL~\cite{rombach2022high} to project both $X_o$ and $X_z$ into a shared latent manifold. Let $\mathcal{E}_{\text{AE}}$ and $\mathcal{D}_{\text{AE}}$ denote the encoder and decoder of the AutoencoderKL, respectively. We define $z_o = \mathcal{E}_{\text{AE}}(x_o)$ and $z_z = \mathcal{E}_{\text{AE}}(x_z)$, where $z_o$ and $z_z$ are the latent embeddings of the observed T2w and DWI modalities, respectively.

Within this latent manifold, we define an optimal transport path between the conditional distributions $p_o(z_o \mid c)$ and $p_z(z_z \mid c)$ given a condition $c$. By setting the condition $c = z_o$, we establish a linear conditional probability path at continuous time $t \in [0, 1]$:
\begin{equation}
p_t(z \mid z_o) = \mathcal{N}\Big(z \mid t \cdot z_z + (1-t) \cdot z_o, \sigma_{\min}^2 I\Big),
\end{equation}
where $\mathcal{N}$ denotes the Gaussian distribution and $I$ is the identity matrix. This linear path enforces intuitive boundary conditions: at $t=0$, the distribution concentrates near $z_o$; at $t=1$, it concentrates near $z_z$, with intermediate states tracing a straight interpolation in the latent space.

The associated vector field $u_t(z_o, z_z) = z_z - z_o$ defines a constant-velocity transition from $z_o$ to $z_z$, with the exact path solution $\phi_t(z_o, z_z) = t \cdot z_z + (1-t) \cdot z_o$. To ensure alignment with this optimal transport trajectory, we minimize the flow matching loss:
\begin{equation}
\mathcal{L}_{\text{FM}} = \mathbb{E}_{t \sim \mathcal{U}(0,1), x_o, x_z} \big\| v_{\theta_G}\big(t, \phi_t(z_o, z_z)\big) - u_t(z_o, z_z) \big\|_2^2,
\end{equation}
where $v_{\theta_G}(\cdot)$ is a parameterized velocity field implemented via a U-Net architecture. Following \cite{schusterbauer2024fmboost}, conditioning the network directly on $z_o$ enforces strong multi-modal coupling, establishing a near one-to-one mapping from the T2w to the DWI latent spaces. In practice, we supervise the velocity field at intermediate states $\phi_t(z_o, z_z)$ to encourage globally consistent transport. Noise augmentation is applied to $z_o$ during training to further regularize the learned field and enhance generation robustness against minor input variations.

\subsection{Cancer Localizer}

The cancer localizer, denoted by $f_L(\cdot;\theta_L)$, predicts an ordinal ISUP \cite{epstein20162014} label for each Barzell zone $r \in R$ from the observed T2w image $x_o$ and the generated DWI $\hat{x}_z$. Since supervision is available only at the zone level rather than as voxel-wise lesion annotations, we formulate this task as zone-level ordinal classification over $C$ groups.

We concatenate $x_o$ and $\hat{x}_z$ and feed them into a 3D ResNet-FPN backbone to obtain multi-scale volumetric features. For each zone $r$, we use its bounding box $b_r$ to extract a fixed-size feature via 3D RoIAlign \cite{he2017mask}. An RoI head then maps the pooled feature to a zone embedding $e_r \in \mathbb{R}^{D_f}$ and zone logits $l_r \in \mathbb{R}^{C}$.

To capture spatial dependencies among nearby zones, we refine the logits with a differentiable conditional random field (CRF) \cite{zheng2015conditional}. We build a symmetrized $k$-nearest-neighbor ($k$-NN) graph from zone centers $s_r$ and define the pairwise affinity as
\begin{equation}
A_{rj}
=
\mathbb{1}\big[(r,j)\in\mathcal{E}_{\mathcal{G}}\big]
\exp\!\left(
-\frac{\|s_r-s_j\|_2^2}{2\sigma_s^2}
\right)
\left(
\frac{1+\cos(e_r,e_j)}{2}
\right),
\end{equation}
where $\sigma_s>0$ is learnable. Row-normalizing $A$ gives propagation weights $W_{rj}=A_{rj}/(\sum_{j'}A_{rj'}+\epsilon)$, where $\epsilon$ is a small constant. We use the classifier logits as unary scores and an ordinal compatibility matrix $\Omega \in \mathbb{R}^{C \times C}$ that penalizes larger label discrepancies more heavily. Mean-field refinement is implemented as
\begin{equation}
q_r^{(\tau+1)}
=
\operatorname{softmax}\!\left(
l_r
-
\lambda \sum_j W_{rj}\, q_j^{(\tau)} \Omega
\right),
\end{equation}
where $\lambda \ge 0$ is learnable and $q_r^{(0)}=\operatorname{softmax}(l_r)$. After $T_{\mathrm{MF}}$ mean-field iterations, the refined class probabilities are taken as $\tilde q_r = q_r^{(T_{\mathrm{MF}})}$.

To respect the ordinal structure of ISUP labels, we optimize the localizer using a class-weighted Earth Mover's Distance (EMD) loss, which penalizes near-miss predictions less severely than standard categorical losses.

\subsection{Generalized Expectation-Maximization}

In this task, DWI images $x_z$ provide valuable diagnostic information for the cancer localizer but are unavailable during inference; instead, they are generated by the latent modality generator $f_G$. We thus treat $x_z$ as a latent variable. Let $\mathcal{T}$ denote the complete training dataset, consisting of: (a) a multi-modality dataset $\mathcal{T}_M = \{(x_o, x_z, y, R)\}$, where paired T2w and DWI images are available alongside zone labels; and (b) a single-modality dataset $\mathcal{T}_S = \{(x_o, y, R)\}$, which lacks $x_z$. For notational brevity, we omit the explicit definitions of the individual data spaces from which the respective random variables are sampled. We model the joint distribution as
\begin{equation}
p(y, x_z \mid x_o, R; \theta) = p\big(y \mid x_o, x_z, R; \theta_L\big)\, p\big(x_z \mid x_o; \theta_G\big),
\label{eq:joint}
\end{equation}
where \(\theta=\{\theta_L, \theta_G\}\) denotes the parameters of the cancer localizer and the latent modality generator. Directly maximizing the marginal likelihood over the latent variable is intractable; therefore, we optimize a variational lower bound that incorporates a KL divergence term. To accomplish this, we develop a Generalized Expectation-Maximization strategy that iteratively improves the lower bound.

\subsubsection*{E-Step}
At iteration \(m\), the posterior over the latent variable \(x_z\) is estimated using the current parameter $\theta^{(m)}$, defining the expected lower bound $Q(\theta \mid \theta^{(m)})$ as
\begin{align}
Q(\theta \mid \theta^{(m)}) =\ &\mathbb{E}_{x_z \sim p(x_z \mid x_o, y, R; \theta^{(m)})} \Big[ \log p(y, x_z \mid x_o, R; \theta) \Big] \notag\\
&- \mathrm{KL}\Big(p(x_z \mid x_o, y, R; \theta^{(m)}) \,\|\, p(x_z \mid x_o; \theta_G)\Big).
\label{eq:lowerbound}
\end{align}
where the expectation $\mathbb{E}_{x_z}[\cdot]$ is taken over the latent variable \(x_z\). It is trivial to note that, in our algorithm, the posterior probability is defined by the downstream cancer localization and requires current estimates of both network parameters. This posterior balances between consistency with the current localizer through the joint log-likelihood term, and adherence to the generator’s conditional prior \(p(x_z \mid x_o; \theta_G)\) through the KL regularizer.

\subsubsection*{M-Step}
Rather than fully maximizing \(Q(\theta \mid \theta^{(m)})\), we employ a GEM update which guarantees a monotonic increase of the lower bound. In practice, we update the parameters via coordinate ascent. First, the modality generator is updated by minimizing
\begin{equation}
\begin{split}
\theta_G^{(m+1)} =\ & \arg\min_{\theta_G}\, \mathbb{E}_{(x_o, x_z, y, R) \sim \mathcal{T}_M} \Big[
L_{\text{FM}}\big(\theta_G; x_o, x_z\big) \\
& + \alpha\, L_{\text{EMD}}\Big(f_L\big(x_o, f_G(x_o; \theta_G); \theta_L^{(m)}\big), y, R\Big)
\Big],
\end{split}
\label{eq:thetaG}
\end{equation}
and then the localizer parameters are refined by minimizing
\begin{align}
\theta_L^{(m+1)}
   &= \arg\min_{\theta_L}\, \mathbb{E}_{(x_o, y, R) \sim \mathcal{T}_S} \Big[ \nonumber \\
   &\quad L_{\text{EMD}}\Big(f_L\big(x_o, f_G(x_o; \theta_G^{(m+1)}); \theta_L\big),
            y, R\Big)\Big].
\label{eq:thetaD}
\end{align}

This GEM-based approach, by alternating updates between \(\theta_G\) and \(\theta_L\), effectively maximizes the variational lower bound and, consequently, approximates the maximization of the marginal likelihood, leading to improved localization performance. It is interesting to discuss two potential variants to the proposed algorithm, by considering prior knowledge over the parameters of the modality generator $\theta_G$ and the latent DWI image itself $x_z$. They lead to two different training strategies, using a fixed, fully pretrained modality generator and using the training-available DWI images (as opposed to the predicted DWI as illustrated in Fig.~\ref{fig2}) as the cancer localizer input, respectively. However, these may be considered beyond the scope of this study.

In our implementation, we first refine the generator so that it both matches observed DWI on paired data and produces latent images that assist the current localizer; we then freeze the generator and update the localizer on all available T2-only samples by using the generator’s outputs as inputs. This alternating coordinate-ascent procedure continuously improves the variational lower bound while avoiding expensive exact maximization at each step. During inference, only \(x_o\) is provided: the generator synthesizes \(\hat{x}_z\), which the localizer uses to predict zone-level cancer probabilities, yielding a T2-only decision pipeline.

\section{Experiments}

\subsection{Datasets} 

\subsubsection{PROMIS Dataset}
The PROMIS dataset \cite{ahmed2017diagnostic} is a publicly available multicenter cohort comprising 3D mpMRI from 566 patients who were blinded to prior biopsy results. The acquisition protocol includes T2w imaging and DWI acquired at \mbox{$b=0$} and high $b$-values (\mbox{$b=1400\,\mathrm{s/mm^2}$}). Histopathological ground truth was established via template-guided prostate mapping biopsies (TPM), evaluated by expert uropathologists. Biopsy cores were localized according to the Barzell zone system. We delineated the spatial location of each Barzell zone based on the prostate gland anatomy following the definitions in \cite{valerio2016transperineal}. Zone-level labels derived from the TPM biopsies provided supervision at the Barzell zone granularity throughout our evaluation. The patient-level pathology distribution is as follows: ISUP 0 (benign, $n=166$), ISUP 1 ($n=98$), ISUP 2 ($n=187$), ISUP 3 ($n=71$), ISUP 4 ($n=20$), and ISUP 5 ($n=24$). At the zone level, the dataset comprises a total of 11,320 zones with the following distribution: ISUP 0 ($n=8,872$), ISUP 1 ($n=1,022$), ISUP 2 ($n=1,012$), ISUP 3 ($n=259$), ISUP 4 ($n=45$), and ISUP 5 ($n=110$). The dataset was subjected to a stratified random partition by patient into three disjoint subsets: 70\% for training, 15\% for testing, and 15\% for validation. The training subset was used to pretrain the Cancer Localizer and to train GEM.

\subsubsection{Targeted Dataset}
The Targeted dataset is a large-scale private cohort collected under ethically approved clinical protocols across multiple institutions in Miami, FL. The patient population consists of individuals with prostate cancer undergoing transperineal interventions, targeted biopsies, or mpMRI targeted focal therapy. These procedures were guided by brachytherapy templates and MRI-transrectal ultrasound (MR-TRUS) fusion systems. This dataset comprises multiparametric 3D MRI scans from 2,091 patients, including T2w sequences and DWI at \mbox{$b=0$} and high $b$-values (\mbox{$b=2000\,\mathrm{s/mm^2}$}). We leveraged this extensive cohort exclusively to pretrain the latent modality generator. This strategy enables the conditional mapping from T2w to DWI to capture a broader spectrum of acquisition protocols and anatomical variability before the generator is fine-tuned via GEM optimization to align with the localization objective and evaluation settings.

\subsubsection{PI-CAI Dataset}
The public Training and Development Dataset of PI-CAI \cite{saha2024artificial} contains 1,500 mpMRI cases from 1,476 patients. At the patient level, the pathological distribution is as follows: ISUP 0 (benign, $n=847$), ISUP 1 ($n=228$), ISUP 2 ($n=234$), ISUP 3 ($n=99$), ISUP 4 ($n=40$), and ISUP 5 ($n=52$). We use this dataset for external patient-level validation.

\subsection{Data Preprocessing}

\begin{figure}[t]
\centering
\includegraphics[width=0.8\columnwidth]{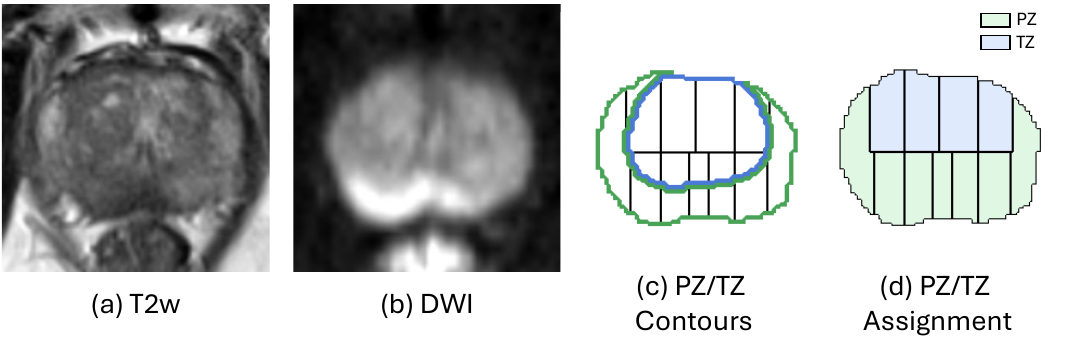}
\caption{Assignment of Barzell zones to the PZ or TZ. (a) T2w. (b) DWI. (c) PZ/TZ contours overlaid on Barzell zones. (d) Final assignment based on the majority overlap rule.} \label{pztz}
\end{figure}

To mitigate spatial misalignment between T2w and DWI sequences, we implemented an anatomy-guided rigid registration using the SimpleITK library. The T2w volume served as the fixed reference, while the low-$b$ DWI (\mbox{$b=0$}) was the moving image. We employed mattes mutual information as the similarity metric. To enhance robustness, the metric computation was restricted to a ROI defined by the prostate gland mask subject to a $5\,\mathrm{mm}$ morphological dilation. The optimized transformation parameters were subsequently applied to the high-$b$ DWI volumes, aligning all modalities to the T2w coordinate space. Following registration, all volumetric data were resampled to a uniform isotropic resolution of $0.5 \times 0.5 \times 0.4 \,\mathrm{mm^3}$. The volumes were center-cropped about the gland centre of mass to a fixed dimension of $128 \times 128 \times 128$ voxels. Finally, linear intensity normalization was applied to scale the voxel intensities to the range $[0,1]$. These steps standardize spatial resolution and intensity distributions across patients from different clinical sites.

\subsection{Evaluation Metrics}

We evaluate the cancer localization performance of the model at three granularities: the Barzell zone\cite{ahmed2017diagnostic}, the region consisting of the peripheral zone (PZ) and transition zone (TZ), and the patient level. The analysis of the TZ and PZ verifies whether the proposed method can compensate for the limitations of T2w imaging, which generally exhibits lower sensitivity in the peripheral zone compared to DWI. As shown in Fig.~\ref{pztz}, the Barzell zones are assigned to the PZ or TZ based on the majority overlap rule. For the region and patient levels, the maximum predicted probability within a specific anatomical area is designated as the risk score. To provide a comprehensive clinical assessment, we report the ISUP group prediction results (ISUP 0, 1, 2, and $\geq$3) and the detection performance across three histopathological malignancy thresholds corresponding to tertiary definition (any cancer, ISUP $\ge$ 1), primary definition (clinically significant prostate cancer, ISUP $\ge$ 2), and secondary definition (aggressive cancer, ISUP $\ge$ 3) \cite{ahmed2017diagnostic}.  Our evaluation framework incorporates standard discriminative metrics including the quadratic weighted kappa (QWK), area under the receiver operating characteristic curve (AUC), macro F1 score, and matthews correlation coefficient (MCC). To evaluate clinical utility, we measure sensitivity at fixed specificities of 80\% and 90\%, alongside specificity at fixed sensitivities of 80\% and 90\%. These operating points are essential for quantifying the clinical tradeoff between minimizing unnecessary biopsies via high specificity and preventing the missed diagnosis of clinically significant lesions via high sensitivity \cite{haj2024magnetic}. Each experiment is repeated three times with different random seeds, and we report the mean and standard deviation.

\subsection{Implementation Details}

We train the AutoencoderKL with a batch size of $2$ and a learning rate of $1\times10^{-4}$ for $50$ epochs on the PROMIS and Targeted. Specifically, the encoder maps the input 3D prostate MRI volume with spatial dimensions $128 \times 128 \times 128$ to a latent representation of shape $3 \times 16 \times 16 \times 16$. The flow-matching model is trained with  a batch size $12$ and a learning rate $1\times10^{-4}$ for $50$ epochs. During inference, we use the exponential moving average (EMA) of the model parameters, with the number of ordinary differential equation sampling steps set to $30$. The cancer localizer is trained with a batch size of $12$ and a learning rate of $1\times10^{-4}$ for $50$ epochs on mpMRI from PROMIS. For CRF refinement, the number of mean-field iterations is set to 1, the number of nearest neighbors is set to 3, and the initial $\sigma$ is set to 5. We initialize both modules with their pre-trained weights and subsequently optimize them jointly under the GEM framework with a batch size of $12$ and a learning rate of $1\times10^{-5}$ for $50$ epochs, with the hyperparameter $\alpha$ set to $0.1$. All models are implemented using PyTorch and MONAI \cite{cardoso2022monai}, and all experiments are performed on 96GB NVIDIA GH200 GPUs in Isambard-AI \cite{mcintosh2024isambard}.

\subsection{Comparison and ablation studies}
To rigorously evaluate the proposed GEM framework, we conducted a series of comparison and ablation experiments. We benchmarked our approach against several reference settings built on the same Cancer Localizer architecture. Specifically, Baseline(T2w), Baseline(DWI) and Baseline(ADC) denote uni-modal baselines trained and evaluated using only T2w, only DWI or only ADC, respectively. Baseline(T2w, DWI), Baseline(T2w, ADC) and Baseline(3M) denote multi-modal upper-bound settings, where the Cancer Localizer is trained and evaluated with two modalities (T2w+DWI or T2w+ADC) or three modalities (T2w+DWI+ADC), respectively. In addition, we compared against adapted state-of-the-art methods for missing modalities (ModDrop \cite{neverova2015moddrop}, SimMLM \cite{li2025simmlm}) and privileged information (KD-Net \cite{hu2020knowledge}). These comparisons were performed at the Barzell zone, region (PZ/TZ) and patient levels under different pathological definitions. Complementing these comparisons, we performed ablation studies to isolate the contributions of key components in the proposed framework, including the the CRF, the modality generator, and the GEM optimization strategy. We also performed an ablation study on the hyperparameter $\alpha$ of GEM.
\section{Results}

\subsection{Comparison Results}

\begin{table*}[htbp]
  \centering
  \caption{Barzell zone-level ISUP group prediction. Best in \textbf{bold}; second best \underline{underlined}.
  }
  \label{tab:zone_metrics_comparison}
  \resizebox{0.73\linewidth}{!}{
    \begin{tabular}{lcccccc}
    \toprule
    Methods & Training&Inference& QWK$\uparrow$ & AUC$\uparrow$ & Macro F1$\uparrow$ & MCC$\uparrow$ \\
    \midrule
    Baseline(T2w,DWI) &T2w,DWI&T2w,DWI & $25.34\pm3.58$ & $68.94\pm0.81$ & $33.68\pm3.35$ & $16.12\pm2.03$ \\
    Baseline(T2w,ADC) &T2w,ADC&T2w,ADC & $10.94\pm0.86$ & $64.47\pm1.50$ & $28.71\pm0.94$ & $12.47\pm2.16$ \\
    Baseline(3M) &T2w,DWI,ADC&T2w,DWI,ADC & $31.36\pm3.42$ & $71.80\pm1.76$ & $35.95\pm3.07$ & $19.02\pm2.85$ \\
    Baseline(DWI)&DWI&DWI & $27.82\pm1.31$ & $69.45\pm0.67$ & $36.29\pm1.41$ & $17.62\pm2.09$ \\
    Baseline(ADC)&ADC&ADC & $20.05\pm3.80$ & $66.56\pm2.78$ & $32.96\pm2.88$ & $16.69\pm1.10$ \\
    \midrule
ModDrop\cite{neverova2015moddrop}  &T2w,DWI&T2w      & $11.42\pm2.02$ & $62.88\pm0.86$ & $27.67\pm1.18$ & $8.21\pm0.94$ \\
SimMLM\cite{li2025simmlm}&T2w,DWI&T2w          & $15.32\pm8.99$ & $66.83\pm3.24$ & $28.75\pm1.88$ & $11.32\pm4.82$ \\
KD-Net \cite{hu2020knowledge}&T2w,DWI&T2w        & $\underline{21.89\pm5.40}$ & $67.11\pm1.65$ & $\underline{33.62\pm3.38}$ & $\underline{16.82\pm3.05}$ \\
Baseline(T2w) &T2w&T2w & $16.72\pm3.59$& $\underline{67.23\pm1.68}$& $30.01\pm1.75$& $13.94\pm0.68$\\
Ours(LDM) &T2w,DWI&T2w      & $21.66\pm3.51$ & $66.04\pm0.88$ & $30.48\pm2.27$ & $15.13\pm1.75$ \\
Ours(LFM) &T2w,DWI&T2w      & $\mathbf{26.68\pm0.65}$ & $\mathbf{70.30\pm2.08}$ & $\mathbf{34.20\pm2.92}$ & $\mathbf{18.06\pm3.14}$ \\
\bottomrule
    \end{tabular}%
    }
\end{table*}

\begin{table*}[htbp]
  \centering
  \caption{Barzell zone-level csPCa detection. Best in \textbf{bold}; second best \underline{underlined}.
  }
  \label{tab:zone_diff_definition}
  \resizebox{\textwidth}{!}{
    \begin{tabular}{lcccccccccccc}
    \toprule
    \multirow{2}{*}{Methods} & \multicolumn{4}{c}{Tertiary Definition} & \multicolumn{4}{c}{Primary Definition} & \multicolumn{4}{c}{Secondary Definition} \\
    \cmidrule(lr){2-5} \cmidrule(lr){6-9} \cmidrule(lr){10-13}
     & \multicolumn{1}{c}{Spe@Sen80\%$\uparrow$} & \multicolumn{1}{c}{Spe@Sen90\%$\uparrow$} & \multicolumn{1}{c}{Sen@Spe80\%$\uparrow$} & \multicolumn{1}{c}{Sen@Spe90\%$\uparrow$} 
     & \multicolumn{1}{c}{Spe@Sen80\%$\uparrow$} & \multicolumn{1}{c}{Spe@Sen90\%$\uparrow$} & \multicolumn{1}{c}{Sen@Spe80\%$\uparrow$} & \multicolumn{1}{c}{Sen@Spe90\%$\uparrow$} 
     & \multicolumn{1}{c}{Spe@Sen80\%$\uparrow$} & \multicolumn{1}{c}{Spe@Sen90\%$\uparrow$} & \multicolumn{1}{c}{Sen@Spe80\%$\uparrow$} & \multicolumn{1}{c}{Sen@Spe90\%$\uparrow$} \\
    \midrule
    Baseline(T2w,DWI)  & $44.74\pm1.49$& $29.58\pm1.76$& $43.74\pm0.79$& $25.98\pm2.23$& $46.73\pm2.08$& $31.03\pm1.65$& $45.87\pm2.34$& $31.35\pm2.23$& $50.95\pm5.65$& $32.93\pm11.05$& $57.69\pm0.00$& $43.59\pm8.67$\\
    Baseline(T2w,ADC)  & $44.82\pm7.23$& $30.24\pm5.99$& $32.60\pm3.43$& $14.43\pm1.24$& $41.23\pm9.66$& $24.22\pm6.43$& $34.19\pm1.44$& $15.68\pm2.73$& $34.71\pm1.66$& $19.62\pm5.21$& $34.62\pm7.69$& $16.03\pm1.11$\\
    Baseline(3M)  & $51.91\pm2.55$& $35.99\pm1.19$& $46.39\pm4.50$& $29.41\pm3.12$& $56.19\pm2.32$& $38.31\pm3.77$& $56.37\pm3.17$& $40.76\pm4.76$& $52.65\pm9.77$& $37.74\pm12.19$& $53.85\pm13.46$& $43.46\pm16.08$\\
    Baseline(DWI) & $45.70\pm2.52$& $30.27\pm2.33$& $42.65\pm1.82$& $28.45\pm0.87$& $49.59\pm2.79$& $33.68\pm2.63$& $51.98\pm1.31$& $38.45\pm1.03$& $48.26\pm4.23$& $33.32\pm6.68$& $55.13\pm2.94$& $40.38\pm13.46$\\
    Baseline(ADC) & $45.70\pm1.07$& $30.87\pm2.21$& $42.65\pm5.63$& $20.82\pm5.12$& $44.70\pm2.39$& $29.22\pm3.05$& $37.72\pm8.46$& $21.09\pm7.19$& $41.24\pm9.98$& $29.82\pm6.73$& $41.67\pm16.58$& $29.49\pm16.36$\\
     \midrule
ModDrop\cite{neverova2015moddrop}         
&$43.90\pm2.98$& $30.65\pm5.68$& $33.52\pm5.38$& $17.63\pm3.48$& $45.86\pm5.15$& $32.02\pm7.55$& $34.65\pm5.51$& $19.31\pm5.51$& $39.24\pm9.76$& $28.12\pm12.55$& $30.13\pm2.22$& $14.74\pm2.94$ \\

SimMLM\cite{li2025simmlm}           
&$43.73\pm6.03$& $30.37\pm3.12$& $37.90\pm7.53$& $21.92\pm7.19$& $46.93\pm8.13$& $30.76\pm5.30$& $41.12\pm7.22$& $22.57\pm6.77$& $\underline{50.53\pm9.62}$& $\mathbf{39.86\pm7.34}$& $46.79\pm8.01$& $26.28\pm4.00$ \\

KD-Net \cite{hu2020knowledge}         
&$45.70\pm1.38$& $31.47\pm1.77$& $41.64\pm2.07$& $22.33\pm0.90$& $41.23\pm2.33$& $25.35\pm3.62$& $44.39\pm1.59$& $\underline{30.99\pm2.78}$& $43.86\pm9.57$& $23.81\pm6.16$& $\underline{49.36\pm8.88}$& $\underline{33.97\pm8.88}$ \\

Baseline(T2w) 
& $\underline{48.21\pm1.98}$& $\underline{34.63\pm2.60}$& $37.99\pm2.33$& $17.95\pm1.09$& $49.54\pm5.34$& $\mathbf{37.59\pm5.02}$& $40.26\pm3.22$& $23.07\pm3.69$& $43.86\pm6.71$& $\underline{33.64\pm9.56}$& $37.82\pm9.87$& $23.72\pm5.88$\\

Ours(LDM)        
&$48.10\pm0.70$& $33.40\pm1.32$& $\mathbf{44.93\pm4.65}$& $\underline{23.97\pm1.55}$& $\underline{52.17\pm1.24}$& $\underline{34.99\pm0.94}$& $\underline{48.02\pm2.80}$& $28.22\pm2.80$& $36.28\pm4.19$& $25.31\pm3.35$& $37.50\pm4.08$& $27.88\pm1.36$ \\

Ours(LFM)       
&$\mathbf{54.30\pm3.23}$& $\mathbf{37.51\pm1.96}$& $\underline{43.29\pm2.32}$& $\mathbf{26.10\pm2.23}$& $\mathbf{55.13\pm3.04}$& $34.47\pm0.81$& $\mathbf{50.74\pm1.05}$& $\mathbf{32.82\pm2.45}$& $\mathbf{53.21\pm4.08}$& $27.32\pm5.00$& $\mathbf{50.00\pm2.72}$& $\mathbf{36.54\pm8.16}$ \\
    \bottomrule
    \end{tabular}%
  }
\end{table*}

\begin{table*}[htbp]
  \centering
  \caption{Region-level (PZ/TZ) csPCa detection under the primary definition. Best in \textbf{bold}; second best \underline{underlined}.}
  \label{tab:pz_tz}
  \resizebox{\textwidth}{!}{
    \begin{tabular}{lcccccccc}
    \toprule
    \multirow{2}{*}{Methods} & \multicolumn{4}{c}{Peripheral Zone} & \multicolumn{4}{c}{Transition Zone} \\
    \cmidrule(lr){2-5} \cmidrule(lr){6-9}
     & \multicolumn{1}{c}{F1$\uparrow$} & \multicolumn{1}{c}{AUC$\uparrow$} & \multicolumn{1}{c}{Spec@Sen80\%$\uparrow$} & \multicolumn{1}{c}{Sen@Spe80\%$\uparrow$} 
     & \multicolumn{1}{c}{F1$\uparrow$} & \multicolumn{1}{c}{AUC$\uparrow$} & \multicolumn{1}{c}{Spec@Sen80\%$\uparrow$} & \multicolumn{1}{c}{Sen@Spe80\%$\uparrow$} \\
    \midrule
    Baseline(T2w,DWI)  & $58.24\pm8.34$& $64.16\pm5.87$& $37.78\pm7.79$& $41.67\pm19.25$& $49.75\pm5.48$& $62.31\pm1.65$& $35.97\pm8.44$& $36.36\pm5.25$ \\
    Baseline(T2w,ADC)  & $64.21\pm1.92$& $64.65\pm8.25$& $47.56\pm18.68$& $27.78\pm12.11$& $54.79\pm6.12$& $59.45\pm4.44$& $43.33\pm9.44$& $30.30\pm3.03$ \\
    Baseline(3M)  & $63.44\pm3.39$& $70.66\pm3.58$& $45.63\pm14.64$& $49.07\pm8.93$& $61.54\pm1.47$& $69.76\pm3.90$& $44.86\pm3.13$& $49.29\pm7.67$ \\
    Baseline(DWI) & $57.17\pm5.29$& $63.33\pm0.33$& $39.11\pm2.70$& $40.74\pm5.78$& $57.82\pm2.89$& $68.54\pm4.71$& $46.25\pm4.91$& $40.61\pm11.70$ \\
    Baseline(ADC) & $62.21\pm1.26$& $69.55\pm3.23$& $56.15\pm1.85$& $39.81\pm10.52$& $59.47\pm2.68$& $61.20\pm4.45$& $39.17\pm10.47$& $31.72\pm2.99$ \\

    \midrule
ModDrop\cite{neverova2015moddrop}         
& $65.89\pm2.20$& $70.35\pm0.46$& $49.48\pm3.78$& $44.44\pm9.62$& $56.21\pm7.08$& $63.40\pm5.43$& $38.47\pm12.32$& $39.80\pm7.91$ \\

SimMLM\cite{li2025simmlm}           
& $63.91\pm5.93$& $66.63\pm7.73$& $41.78\pm10.36$& $45.37\pm19.71$& $59.38\pm3.94$& $66.60\pm4.79$& $\underline{49.44\pm3.76}$& $41.01\pm6.54$ \\

KD-Net \cite{hu2020knowledge}         
& $63.36\pm5.73$& $66.56\pm1.10$& $35.11\pm4.00$& $47.22\pm7.35$& $55.01\pm6.93$& $65.97\pm1.64$& $42.64\pm9.38$& $\underline{44.44\pm0.93}$ \\

Baseline(T2w) 
& $63.15\pm1.49$& $70.76\pm2.00$& $58.52\pm7.56$& $\underline{51.85\pm3.21}$& $54.34\pm5.30$& $67.59\pm2.92$& $45.97\pm4.28$& $43.84\pm10.59$\\

Ours(LDM)      
& $\underline{71.96\pm0.76}$& $\mathbf{74.85\pm0.31}$& $\mathbf{65.33\pm1.89}$& $\mathbf{65.28\pm9.82}$& $\underline{62.59\pm0.46}$& $\underline{67.68\pm0.27}$& $\mathbf{50.42\pm0.00}$& $36.36\pm12.86$ \\

Ours(LFM)       
& $\mathbf{73.57\pm1.20}$& $\underline{74.07\pm0.26}$& $\underline{65.11\pm3.46}$& $44.44\pm7.86$& $\mathbf{63.95\pm2.76}$& $\mathbf{67.90\pm8.44}$& $40.62\pm13.26$& $\mathbf{46.36\pm7.29}$ \\
\bottomrule
    \end{tabular}%
  }
\end{table*}

We validated the efficacy of the proposed GEM framework against standard baselines, state-of-the-art missing modality and LUPI methods, and practical upper bounds. Quantitative results (Table~\ref{tab:zone_metrics_comparison}, Table~\ref{tab:zone_diff_definition}, Table~\ref{tab:pz_tz} and Table~\ref{tab:patient}) demonstrate the clinical and methodological advantages of our approach.

\subsubsection{Comparative Analysis of Modality-Specific Baselines}
Comparative analysis of the experimental baselines highlights the role/importance of high b-value DWI over ADC for prostate cancer localization. At the zone level, the uni-modal Baseline(DWI) achieves a QWK of $27.82 \pm 1.31$ and an AUC of $69.45 \pm 0.67$, substantially outperforming the Baseline(ADC) which only reaches $20.05 \pm 3.80$ and $66.56 \pm 2.78$, respectively. This justifies the selection of high b-value DWI as the privileged latent modality, as the raw signal retains morphological and functional contrasts. Furthermore, while the integration of all three modalities in the Baseline(3M) configuration establishes the theoretical upper bound with a QWK of $31.36 \pm 3.42$, restricting the multi-modal fusion to only two inputs reveals a stark representational conflict. Specifically, Baseline(T2w,DWI) maintains a robust QWK of $25.34 \pm 3.58$, whereas substituting the diffusion signal with the ADC in Baseline(T2w,ADC) causes a performance collapse to $10.94 \pm 0.86$, falling even below the Baseline(T2w) of $16.72 \pm 3.59$. Ultimately, although the standalone T2w inference demonstrates a moderate baseline capacity with a zone-level AUC of $67.23 \pm 1.68$, its lower F1 score of $30.01 \pm 1.75$ compared to diffusion-inclusive models underscores the inherent limitations of relying solely on anatomical sequences, thereby supporting our proposed generative approach that adds learned functional prior from latent diffusion images.

\begin{table*}[htbp]
  \centering
  \caption{Patient-level csPCa detection under the primary definition. Best in \textbf{bold}; second best \underline{underlined}.}
  \label{tab:patient}
  \resizebox{\textwidth}{!}{
    \begin{tabular}{lcccccccc}
    \toprule
    \multirow{2}{*}{Methods} & \multicolumn{4}{c}{PROMIS} & \multicolumn{4}{c}{PI-CAI} \\
    \cmidrule(lr){2-5} \cmidrule(lr){6-9}
     & \multicolumn{1}{c}{F1$\uparrow$} & \multicolumn{1}{c}{AUC$\uparrow$} & \multicolumn{1}{c}{Spec@Sen80\%$\uparrow$} & \multicolumn{1}{c}{Sen@Spe80\%$\uparrow$} 
     & \multicolumn{1}{c}{F1$\uparrow$} & \multicolumn{1}{c}{AUC$\uparrow$} & \multicolumn{1}{c}{Spec@Sen80\%$\uparrow$} & \multicolumn{1}{c}{Sen@Spe80\%$\uparrow$} \\
    \midrule
    Baseline(T2w,DWI)  & $65.39\pm3.94$& $64.06\pm1.94$& $33.33\pm6.78$& $38.81\pm9.83$& $51.15\pm1.60$& $71.84\pm1.56$& $50.67\pm1.88$& $48.25\pm2.89$ \\
    Baseline(T2w,ADC)  & $72.18\pm1.90$& $62.41\pm4.04$& $42.74\pm7.83$& $27.41\pm7.33$& $45.49\pm0.41$& $63.58\pm2.09$& $36.66\pm2.72$& $35.06\pm4.72$ \\
    Baseline(3M)  & $68.98\pm2.28$& $69.67\pm5.37$& $49.57\pm7.40$& $46.67\pm13.33$& $49.99\pm1.08$& $72.34\pm1.38$& $51.05\pm2.76$& $50.90\pm3.50$ \\
    Baseline(DWI) & $67.52\pm5.93$& $62.94\pm0.34$& $35.90\pm2.56$& $39.56\pm5.98$& $49.29\pm3.92$& $69.93\pm2.18$& $45.09\pm3.11$& $46.29\pm2.36$ \\
    Baseline(ADC) & $72.71\pm2.85$& $66.12\pm3.44$& $44.44\pm7.40$& $32.00\pm7.75$& $46.08\pm0.15$& $64.51\pm1.74$& $38.56\pm3.01$& $36.12\pm2.13$ \\

    \midrule
ModDrop\cite{neverova2015moddrop}         & $66.71\pm4.75$& $66.19\pm3.09$& $45.30\pm8.24$& $\underline{41.78\pm4.24}$& $46.70\pm1.09$& $62.55\pm2.77$& $36.33\pm2.59$& $34.64\pm6.31$ \\
SimMLM\cite{li2025simmlm}          & $67.67\pm4.01$& $66.00\pm5.76$& $43.59\pm6.78$& $40.30\pm15.05$& $47.14\pm2.67$& $63.88\pm2.83$& $\underline{37.70\pm5.65}$& $36.84\pm3.72$ \\
KD-Net \cite{hu2020knowledge}         & $62.82\pm6.44$& $63.10\pm0.87$& $28.21\pm5.13$& $40.44\pm5.13$& $43.83\pm0.86$& $58.65\pm1.62$& $29.30\pm4.31$& $31.07\pm1.83$ \\
Baseline(T2w) & $67.69\pm2.87$& $68.74\pm2.72$& $47.01\pm8.24$& $\mathbf{46.37\pm11.90}$& $\underline{47.34\pm1.58}$& $\underline{64.71\pm2.47}$& $36.11\pm6.11$& $\underline{39.07\pm3.45}$\\
Ours(LDM)        & $\underline{70.77\pm1.84}$& $\underline{70.14\pm1.61}$& $\underline{51.28\pm7.25}$& $39.78\pm3.46$& $47.13\pm1.40$& $64.41\pm1.62$& $37.53\pm2.07$& $\mathbf{39.72\pm4.54}$ \\
Ours(LFM)       & $\mathbf{74.83\pm0.86}$& $\mathbf{70.51\pm0.77}$& $\mathbf{55.13\pm5.44}$& $39.56\pm6.91$& $\mathbf{48.26\pm1.01}$& $\mathbf{65.40\pm0.32}$& $\mathbf{40.60\pm1.45}$& $38.00\pm1.43$ \\
\bottomrule
    \end{tabular}%
  }
\end{table*}

\subsubsection{Zone-Level ISUP group And Clinically Significant Cancer Detection} The evaluation at the Barzell zone validates the effectiveness of the proposed framework for T2-Only inference. Our flow matching-based approach, Ours(LFM), improves upon Baseline(T2w) in ISUP group prediction, increasing the QWK from $16.72 \pm 3.59$ to $26.68 \pm 0.65$ and the AUC from $67.23 \pm 1.68$ to $70.30 \pm 2.08$. Compared to other missing-modality techniques, our approach outperforms ModDrop, SimMLM, and KD-Net across all ordinal metrics. While KD-Net achieves a moderate QWK of $21.89 \pm 5.40$, our method yields lower variance and a superior MCC of $18.06 \pm 3.14$, indicating a more stable and balanced ordinal prediction. Notably, Ours(LFM) also surpasses the true multi-modal Baseline(T2w,DWI) which scores a QWK of $25.34 \pm 3.58$. Under the primary definition for clinically significant cancer detection, Ours(LFM) increases the specificity at a fixed 80\% sensitivity to $55.13 \pm 3.04$, outperforming Baseline(T2w) at $49.54 \pm 5.34$ and closely approaching the multi-modal Baseline(3M) upper bound of $56.19 \pm 2.32$. Similarly, under the secondary definition targeting higher grade cancers, our method achieves a specificity of $53.21 \pm 4.08$ at 80\% sensitivity, contrasting with the Baseline(T2w) performance of $43.86 \pm 6.71$.

\subsubsection{Bridging Diagnostic Gaps In Peripheral And Transition Zones} In the peripheral zone, the limited diagnostic specificity of T2w images renders T2-only inference particularly challenging; consequently, PI-RADS dictates the use of diffusion-weighted sequences as the primary modality. Region-level analysis shows that by injecting latent functional information, Ours(LFM) successfully compensates for the modality deficit, achieving a peripheral zone F1 score of $73.57 \pm 1.20$. This surpasses not only the Baseline(T2w) score of $63.15 \pm 1.49$ but also the multi-modal Baseline(3M) score of $63.44 \pm 3.39$. Similarly, the transition zone poses a diagnostic challenge, for example, due to common heterogeneous benign hyperplasia. Here, our method increases the F1 score from $54.34 \pm 5.30$ to $63.95 \pm 2.76$ compared to Baseline(T2w). This localization advantage implies that the probabilistic generation process acts as an effective regularizer, distilling tumor-specific functional characteristics while filtering out structural noise or modality-specific artifacts that frequently degrade actual multi-parametric inputs.

\subsubsection{Patient-Level Risk Stratification}

Patient-level risk stratification confirms the clinical viability of the model. On the internal PROMIS dataset under the primary definition, our method achieves an exceptional F1 score of $74.83 \pm 0.86$ and an area under the curve of $70.51 \pm 0.77$, strictly dominating the uni-modal Baseline(T2w) and surprisingly outperforming the multi-modal Baseline(3M) configuration. This confirms that marginalizing over latent modality uncertainty effectively prevents the model from overfitting to localized artifacts. Conversely, external validation on the PI-CAI dataset evaluates the generalizability of generative modality completion across diverse clinical centers. Specifically, the mean F1 score and AUC of our method ($48.26 \pm 1.01$ and $65.40 \pm 0.32$, respectively) are higher than those of Baseline(T2w) ($47.34 \pm 1.58$ and $64.71 \pm 2.47$). However, this performance remains lower than that of the actual Baseline(T2w,DWI), which achieves an AUC of $71.84 \pm 1.56$. We attribute this gap to a domain shift affecting the generative module, as the AutoencoderKL was trained exclusively on the PROMIS and Targeted cohorts and not on PI-CAI. Consequently, there is a substantial degradation in T2w reconstruction fidelity when transferring from PROMIS to PI-CAI, evidenced by noticeable shifts in MAE ($1.01 \times 10^{-2} \rightarrow 2.05 \times 10^{-2}$), MSE ($1.91 \times 10^{-4} \rightarrow 8.89 \times 10^{-4}$), and PSNR ($37.55 \rightarrow 31.78$). 

\subsection{Ablation Study}

\begin{table}[htbp]
  \centering
  \caption{Ablation study of different components on ISUP group prediction performance.}
  \label{tab:ablation_study}
\resizebox{\linewidth}{!}{
  \begin{tabular}{ccccccc}    \toprule
       CRF& Modality Generator & GEM & QWK$\uparrow$ & AUC$\uparrow$ & Macro F1$\uparrow$ & MCC$\uparrow$ \\
    \midrule
     & & & $9.21\pm2.21$& $59.74\pm1.67$& $25.56\pm3.91$& $9.06\pm3.13$ \\
     \checkmark & & & $9.54\pm6.17$ & $60.75\pm1.13$ & $25.88\pm3.55$ & $9.90\pm5.06$ \\
      & \checkmark & & $17.55\pm2.94$ & $66.35\pm0.82$ & $30.87\pm3.17$ & $13.53\pm1.85$ \\
     \checkmark & \checkmark & & $19.99\pm3.78$ & $66.21\pm1.72$ & $31.82\pm1.99$ & $14.59\pm2.14  $ \\
     \checkmark & \checkmark & \checkmark & $\mathbf{26.68\pm0.65}$ & $\mathbf{70.30\pm2.08}$ & $\mathbf{34.20\pm2.92}$ & $\mathbf{18.06\pm3.14}$ \\

    \bottomrule
    \end{tabular} 
    }
\end{table}

\subsubsection{Impact of Modality Substitution}The baseline model, evaluated using T2w images and zero-filled arrays to simulate missing diffusion sequences, yielded a QWK of $9.21 \pm 2.21$ and an AUC of $59.74 \pm 1.67$. Substituting the zero-filled arrays with synthetic DWI images generated by the modality generator increased the QWK to $17.55 \pm 2.94$. Because this generation step was optimized exclusively for image reconstruction and decoupled from the classification objective, the resulting representations demonstrated restricted diagnostic utility for target localization.

\subsubsection{Efficacy of Spatial Regularization} The integration of a CRF exhibited a dependency on input feature quality. Applied to the zero-filled baseline, the CRF provided minimal metric changes (QWK: $9.54 \pm 6.17$) and increased prediction variance. Conversely, combining the CRF with the decoupled modality generator increased the QWK to $19.99 \pm 3.78$ and the MCC to $14.59 \pm 2.14$. These data indicate that spatial relational smoothing between anatomical zones requires informative multi-modal representations to function effectively.

\begin{figure*}
\centering
\includegraphics[width=\linewidth]{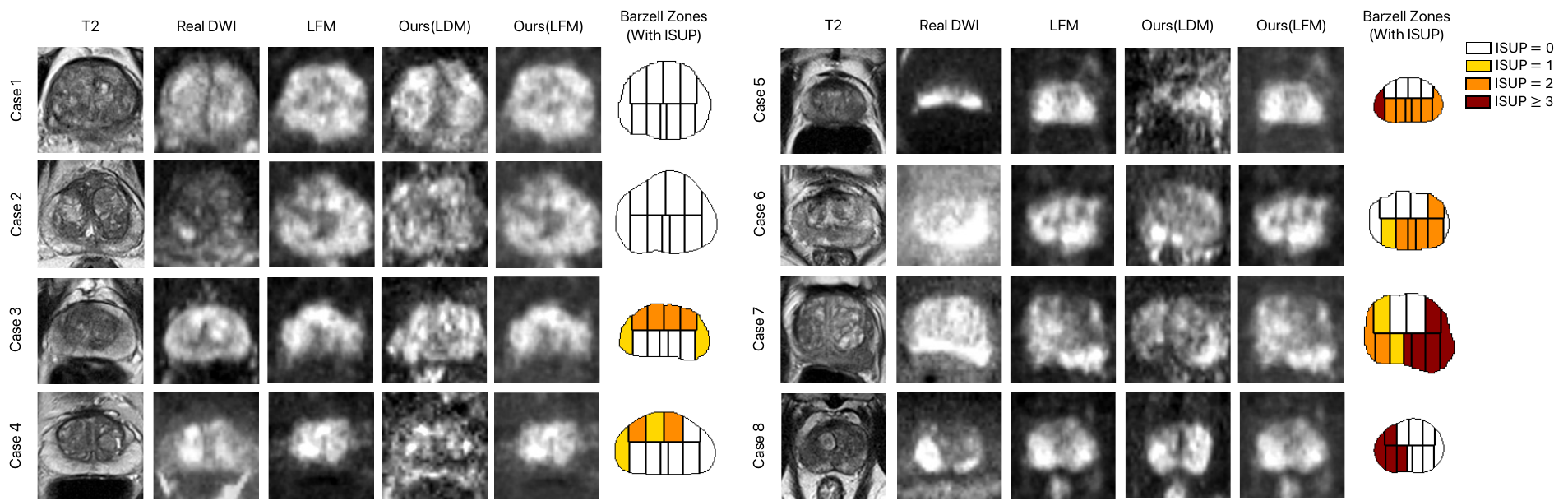}  
\caption{Samples of T2w images, corresponding real DWI, and DWI synthesized by pretrained LFM, Ours(LDM), and Ours(LFM). The rightmost column shows the corresponding Barzell zone labels by ISUP grade, where white denotes ISUP$=$0, yellow denotes ISUP$=$1, orange denotes ISUP$=$2, and red denotes ISUP$\ge$3.}
\label{fig3}
\end{figure*}

\subsubsection{Joint Optimization via Generalized Expectation-Maximization} Concurrent integration of the CRF, the modality generator, and the GEM strategy achieved the highest overall performance. This joint optimization yielded a QWK of $26.68 \pm 0.65$, an AUC of $70.30 \pm 2.08$, and a Macro F1 of $34.20 \pm 2.92$. The generator is optimized to hallucinate diffusion characteristics that explicitly maximize the expected likelihood of accurate cancer localization. The generalized expectation-maximization framework effectively harmonizes the generative and discriminative objectives.

\subsubsection{Sensitivity to the Classification Loss Weight}

\begin{table}[htbp]
  \centering
  \caption{Ablation study of different $\alpha$ on ISUP group prediction performance.}
  \label{tab:ablation_study_alpha}
\resizebox{1.0\linewidth}{!}{
  \begin{tabular}{ccccc}    \toprule
     $\alpha$ & QWK$\uparrow$ & AUC$\uparrow$ & Macro F1$\uparrow$ & MCC$\uparrow$ \\
    \midrule
     0.05 & $24.67\pm1.62$ & $68.84\pm0.17$ & $33.69\pm1.12$ & $17.68\pm0.28$ \\
     0.1 & $\mathbf{26.68\pm0.65}$ & $\mathbf{70.30\pm2.08}$ & $\mathbf{34.20\pm2.92}$ & $\mathbf{18.06\pm3.14}$ \\
     0.2 & $24.44\pm0.94$& $68.83\pm0.21$& $33.82\pm1.23$& $17.80\pm0.25$ \\
     0.5 & $24.11\pm0.67$& $68.84\pm0.22$& $33.52\pm0.91$& $17.29\pm0.66$ \\
    \bottomrule
    \end{tabular}
    }
\end{table}

The diagnostic performance of the framework depended on the hyperparameter $\alpha$, which determines the classification loss weight backpropagated into the generative module. As shown in Table~\ref{tab:ablation_study_alpha},  optimal performance was observed at $\alpha = 0.1$, yielding an MCC of $18.06 \pm 3.14$. Decreasing $\alpha$ to $0.05$ reduced the QWK to $24.67 \pm 1.62$. Increasing $\alpha$ beyond the optimal value caused a progressive decline, with the QWK decreasing to $24.44 \pm 0.94$ at $\alpha = 0.2$ and $24.11 \pm 0.67$ at $\alpha = 0.5$. This suggests that disproportionate discriminative gradients may distort the generative trajectory and impair functional feature synthesis.

\subsection{Qualitative Analysis and Visualization}

Fig.~\ref{fig3} demonstrates the qualitative advantages of the proposed framework across diverse pathological states and anatomical zones. The synthesized latent modalities effectively capture task-relevant functional contrasts, successfully delineating clinically significant prostate cancers as distinct focal hyperintensities (Cases 3--8, which harbor ISUP $\ge$ 2 lesions) while avoiding false-positive signals in benign prostates (Cases 1--2, ISUP 0). Notably, the generated DWI enhances cancer conspicuity in both the peripheral zone (PZ; e.g., Cases 5--6) and the transition zone (TZ; e.g., Cases 3--4). This effectively compensates for the inherent lack of T2w diagnostic specificity in the PZ and mitigates the confounding effects of heterogeneous benign prostatic hyperplasia (BPH) that frequently mimics malignancies in the TZ. Crucially, while real DWI is routinely compromised by severe acquisition-related degradations---such as rectal gas-induced susceptibility artifacts, low signal-to-noise ratios, and severe geometric distortions(Case 5--6)---our T2w-conditioned generative approach inherently bypasses these physical limitations. By leveraging the high spatial resolution of the T2w structural prior, it yields visually artifact-free functional representations that closely align with true anatomical contours. When comparing generative backbones, the flow matching-based Ours(LFM) consistently suppresses the granular noise and structural deformations frequently introduced by the diffusion-based Ours(LDM), resulting in smoother and more anatomically faithful syntheses. Furthermore, the macroscopic visual appearance of the standalone generator (LFM) and the GEM-optimized outputs (Ours(LFM)) is largely similar, whereas Ours(LFM) achieves clear quantitative gains in diagnostic performance. This finding is in line with the design of GEM joint optimization, which uses a small learning rate and EMA weights to adjust pathology-related latent representations while maintaining the stability of the pre-trained generative model. Accordingly, the improvement is more likely attributable to enhanced task-relevant feature representation than to obvious morphological changes at the image level. Additionally, LFM shows higher efficiency, generating a $128 \times 128 \times 128$ DWI volume in 8 seconds versus LDM's 10 seconds.

\section{Conclusion}
This work introduces a novel Generalized Expectation-Maximization framework that enables accurate localization of zonal prostate cancer using only T2-weighted MRI data during inference. By incorporating DWI information during training and jointly optimizing both the latent modality generator and the cancer localization network, our approach outperforms all evaluated baselines. Furthermore, the proposed flow matching model demonstrates significant advantages over existing methods in both computational efficiency and image quality. These findings highlight the feasibility and effectiveness of a diagnostic pipeline that relies solely on T2-weighted images, offering a promising solution to enhance the accessibility and efficiency of prostate cancer care pathway.

\bibliographystyle{IEEEtran}
\bibliography{references}

\end{document}